\documentclass[12pt]{article}
\usepackage{authblk}
\usepackage{geometry}
\usepackage{amssymb}
\usepackage{amsmath}
\usepackage{mathrsfs}
\usepackage[dvips]{graphicx}
\usepackage[dvips]{epsfig}
\usepackage{array}
\usepackage{url}
\usepackage{color}
\usepackage{tikz}
\usepackage{pgf}
\usepackage{tabularx,ragged2e,booktabs,caption}
\usepackage{pdfpages}
\usepackage{forest}
\usepackage{algpseudocode,algorithm,algorithmicx}

\usepackage{hyperref}
\newcommand*\Let[2]{\State #1 $\gets$ #2}
\algrenewcommand\algorithmicrequire{\textbf{Precondition:}}
\algrenewcommand\algorithmicensure{\textbf{Postcondition:}}

\newcommand{\argmin}{\operatornamewithlimits{argmin}}
\newcommand{\argmax}{\operatornamewithlimits{argmax}}

\newcolumntype{C}[1]{>{\Centering}m{#1}}

\usetikzlibrary{intersections}
\usetikzlibrary{arrows}
\usetikzlibrary{shapes, automata, decorations.markings}
\usetikzlibrary{arrows,decorations.pathmorphing,backgrounds,positioning,fit,matrix}

\pgfdeclarelayer{nodelayer}
\pgfdeclarelayer{edgelayer}
\pgfsetlayers{edgelayer,nodelayer,main}

\author[1]{B Ravi Kiran} 
\author[2]{Jean Serra}
\affil[1]{CRIStAL Lab, UMR 9189, Universit\'e Charles de Gaulle, Lille 3 \url{kiran.ravi@univ-lille3.fr}}
\affil[2]{Universit\'e Paris-Est, A3SI-ESIEE LIGM \url{jean.serra@esiee.fr}}

\title{Cost-complexity pruning of random forests}

\begin{document}

\maketitle              

\begin{abstract}
Random forests perform boostrap-aggregation by sampling the training samples with replacement.
This enables the evaluation of out-of-bag error which serves as a internal cross-validation mechanism.
Our motivation lies in using the unsampled training samples to improve each decision tree in the ensemble.
We study the effect of using the out-of-bag samples to improve the generalization error 
first of the decision trees and second the random forest by post-pruning. A prelimiary empirical study on
four UCI repository datasets show consistent decrease in the size of the forests without considerable loss in accuracy.
\footnote{Previous version in proceedings ISMM 2017.}
\end{abstract}
\textbf{Keywords}:Random Forests, Cost-complexity Pruning, Out-of-bag

\section{Introduction}
Random Forests \cite{breiman2001randomForests} is an ensemble method which 
predicts by averaging over multiple instances of classifiers/regressors created by
randomized feature selection and bootstrap aggregation (Bagging). The model
is one of the most consistently performing predictor in many real world
applications \cite{Criminisi_RF_book2012}. Random forests use CART decision 
tree classifiers \cite{BreimanbookCART1984} as weak learners. Random forests
combine two methods : Bootstrap aggregation \cite{breiman1996bagging} (subsampling 
input samples with replacement) and Random subspace \cite{ho1998randomsubspace} 
(subsampling the variables without replacement). There has been continued work during 
the last decade on new randomized ensemble of trees. Extremely randomized
trees \cite{Geurts2006ExtraTrees} where instead of choosing the best split
among a subset of variables under search for maximum information gain, a random
split is chosen. This improves the prediction accuracy. In furthering the understanding of 
random forests \cite{denil_nando_2014_RF} split the training set points into structure 
points: which decide split points but are not involved in prediction, estimation points: 
which are used for estimation. The partition into two sets are done randomly to keep 
consistency of the classifier.  

Over-fitting occurs when the statistical model fits noise or misleading points
in the input distribution, leading to poor generalization error and performance.
In individual decision tree classifiers grown deep, until each input sample can 
be fit into a leaf, the predictions generalizes poorly on unseen data-points. 
To handle this decision trees are pruned. There has been a decade of study on the different pruning methods, 
error functions and measures \cite{RakotomalalaThesis_pruning_1997}, \cite{torgo1999inductive}. 
The common procedure follow is : 1. Generate a set in "interesting trees", 2. Estimate the true 
performance of each of these trees, 3. Choose the best tree. This is called post-pruning
since we grow complete decision trees and then generate a set of interesting trees.
CART uses cost-complexity pruning by associating with each cost-complexity parameter a nested subtree \cite{ESL_2009}.

Though there has been extensive study on the different error functions to perform post-pruning 
\cite{mingers1989empirical}, \cite{weiss1994decision}, there have been very few 
studies performed on pruning random forests and tree ensembles. In practice Random forests
are quite stable with respect to parameter of number of tree estimators. They are shown 
to converge asymptotically to the true mean value of the distribution. \cite{ESL_2009} (page 596)
perform an elementary study to show the effect tree size on prediction performance by fixing 
minimum node size (smaller it is the deeper the tree). This choice of the minimum node size 
are difficult to justify in practice for a given application. Furthermore \cite{ESL_2009} discuss that 
rarity of over-fitting in random forests is a claim, and state that this asymptotic limit can 
over-fit the dataset; the average of fully grown trees can result in too rich a model, and incur unnecessary
variance. \cite{segal2004machine} demonstrates small gains in performance by
controlling the depths of the individual trees grown in random forest.

Finally random forests and tree ensembles are generated by multiple randomization methods.
There is no optimization of an explicit loss functions. The core principle in these methods 
might be interpolation, as shown in this excellent study \cite{wyner2015explaining}. 
Though another important principle is the use of non-parametric density estimation
in these recursive procedures \cite{BiauDevroyeLugosi_Conistency_RF_2008}.

In this paper we are primarily motivated by the internal cross-validation mechanism 
of random forests. The out-of-the-bag (OOB) samples are the set of data points that were 
not sampled during the creation of the bootstrap samples to build the individual trees. 
Our main contribution is the evaluation of the predictive performance cost-complexity pruning 
on random forest and other tree ensembles under two scenarios :
1. Setting the cost-complexity parameter by minimizing the individual tree prediction error on OOB samples for each tree.
2. Setting the cost-complexity parameter by minimizing average OOB prediction error by the forest on all training samples.

In this paper we do not study ensemble pruning, where the idea is to prune complete instances of 
decision trees away if they do not improve the accuracy on unseen data.

\subsection{Notation and Formulation}
Let $Z = \{\mathbf{x}^i, y^i\}_N$ be set of $N$ (input, output) pairs
to be used in the creation of a predictor. Supervised learning consists of two 
types of prediction tasks : regression and classification problem, where in 
the former we predict continuous target variables, while in the latter we predictor
categorical variables. The inputs are assumed to belong to space 
$X := \mathbb{R}^d$ while $Y := \mathbb{R}$ for regression and $Y := \{C_i\}_K$
with $K$ different abstract classes. A supervised learning problem aims to 
infer the function $f : X \to Y$ using the empirical samples $Z$ that 
``generalizes'' well. 

Decision trees fundamentally perform data adaptive non-parametric density estimation
to achieve classification and regression tasks. Decision trees evaluate the density function 
of the joint distribution $P(X, Y)$ by recursively splitting the feature space $X$ greedily, 
such that after each split or subspace, the $Y$s in the children become ``concentrated'' or
in some sense well partitioned. The best split is chosen by evaluating the information gain(chage in entropy)
produced before and after a split. Finally at the leaves of the decision trees one is able to evaluate
the class/value by observing the subspace (like the bin for histograms) and predicting
the majority class respectively \cite{probtheory_PR_1996}.

Given a classification target variable with $C_k = \{1, 2, 3, ... K\}$ classes, 
we denote the proportion of of class $k$ in node  as :

\begin{equation}
	\hat{p}_{tk} = \frac{1}{n_t} \sum_{\mathbf{x}_i \in R_t} I(y_i = k) 
\end{equation}
 
which represents proportion of classifications in node $t$ in decision region $R_t$ with $n_t$ observations.
The prediction in case of classification is performed by taking the majority vote in a leaf, 
i.e. $\hat{y}_i = \argmax_{k} \hat{p}_{tk}$. The misclassification error is given by :

\begin{equation}
l(y,\hat{y}) = \frac{1}{N_t} \sum_{i \in R_t} I(y_i \neq \hat{y}_i) = 1-\hat{p}_{mt}
\end{equation}

The general idea in classification trees is the recursive partition of $\mathbb{R}^d$
by axis parallel splits while maximizing the gini coefficient : 
\begin{equation}
\sum_{k\neq k^\prime} \hat{p}_{mt} \hat{p}_{mt^{\prime}} = 
\sum_{k=1}^K \hat{p}_{mt} (1-\hat{p}_{mt})
\end{equation}

In a decision split the parameters are the split dimension denoted by $j$
and the split threshold $c$. Given an input decision region $S$ we are 
looking for the dimension (here in three dimensions) that minimizes entropy. 
Since we are splitting along $d$ unordered variables, there are $2^{d-1}-1$ possible 
partitions of the $d$ values into two groups (splits) and is computationally prohibitive.
We greedily decide the best split on a subset of variables. We apply this procedure iteratively 
till the termination condition.

\begin{figure}[htbp]
	\centering
	\includegraphics[width=0.95\textwidth]{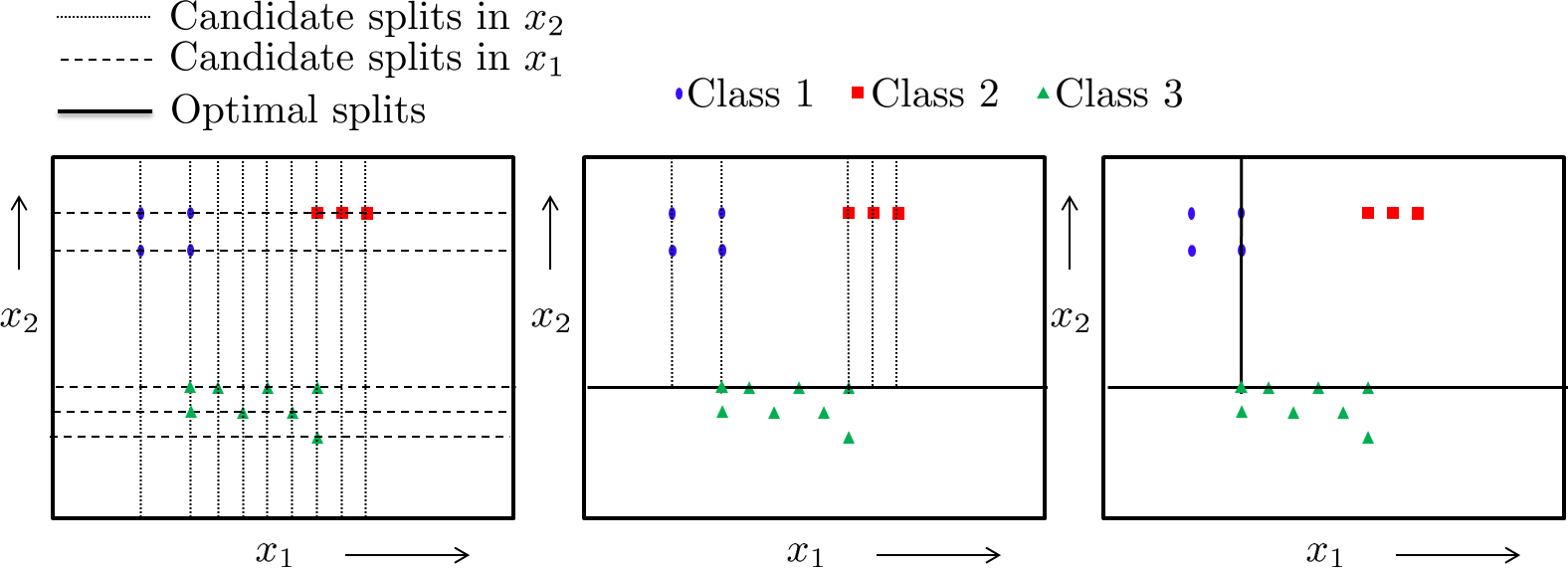}
	\caption{Choosing the axis and splits. Given $\mathbb{R}^2$ feature space,
	we need to chose the best split given we chose a single axis. This consists in 
	choosing all the coordinates along the axis, at which there are points. The 
	optimal split is one that separates the classes the best, according to the
	impurity measure, entropy or other splitting measures. Here we show the 
	sequence of two splits. Since there are finite number of points and feature pairs, 
	there are finite number of splits possible.}
	\label{fig:decision_split_choice}
\end{figure}

As shown in figure \ref{fig:decision_split_choice} the set of splits over which 
the splitting measure is minimized is determined by the coordinates of the training set points.
The number of variables or dimension $d$ can be very large (100s-1000 in bio-informatics). 
Most frequently in CART one considers the sorted coordinates
and from them the split points where the class $y$ change and finally one picks the split that minimizes 
the purity measure best. 

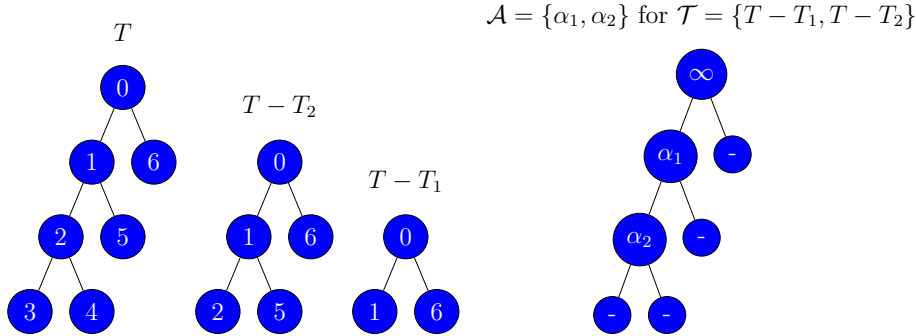
\begin{figure}
	\centering
	\scalebox{0.8}{
	\begin{forest}
		for tree={circle,draw, fill=blue, text=white}
		[0 
		  [1
		    [2 [3][4]]
		    [5] ]
		  [6]
		]
		\node[above=30pt,align=center,anchor=center] {$T$};
	\end{forest}}
	\scalebox{0.8}{
	\begin{forest}
		for tree={circle,draw, fill=blue, text=white}
		[0 
		  [1
		    [2]
		    [5] ]
		  [6]
		]
		\node[above=30pt,align=center,anchor=center] {$T-T_2$};
	\end{forest}}
	\scalebox{0.8}{
	\begin{forest}
		for tree={circle,draw, fill=blue, text=white}
		[0 
		  [1]
		  [6]
		]
		\node[above=30pt,align=center,anchor=center] {$T-T_1$};
	\end{forest}}
	\scalebox{0.8}{
	\begin{forest}
		for tree={circle,draw, fill=blue, text=white}
		[$\infty$ 
		  [$\alpha_1$
		    [$\alpha_2$ [-][-]]
		    [-] ]
		  [-]
		]
		\node[above=30pt,align=center,anchor=center] {$\mathcal{A} = \{\alpha_1, \alpha_2\}$ for $\mathcal{T} = \{T-T_1, T-T_2\}$};
	\end{forest}}
	\caption{Figure shows shows a sequence of nested subtrees $\mathcal{T}$ and the values of cost-complexity 
		parameters associated with these subtrees $\mathcal{A}$ calculated by equation (\ref{eqn:cost-complexity-param})
		and algorithm (\ref{alg:pruning_trees_cost_complexity}). Here $\alpha_2 < \alpha_1 \implies T-T_1 \subset T-T_2$}
	\label{fig:cost_complexity_example}
\end{figure}

\subsection{Cost-Complexity Pruning}

The decision splits near the leaves often provide pure nodes with very narrow decision regions that are
over-fitting to a small set of points. This over-fitting problem is resolved in 
decision trees by performing pruning \cite{BreimanbookCART1984}. There are 
several ways to perform pruning : we study the cost-complexity pruning here.
Pruning is usually not performed in decision tree ensembles, for example
in random forest since bagging takes care of the variance produced by unstable decision trees.
Random subspace produces decorrelated decision tree predictions, which explore different
sets of predictor/feature interactions. 

The basic idea of cost-complexity pruning is to calculate a cost function 
for each internal node. An internal node is all nodes that are not the leaves
nor the root node in a tree. The cost function is given by \cite{ESL_2009}:

\begin{equation}
	R_\alpha (T) = R(T) + \alpha \cdot |\text{Leaves}(T)|
\end{equation}

where 
\begin{equation}
	R(T) = \sum_{t \in \text{Leaves}(T)} r(t) \cdot p(t)  = \sum_{t \in \text{Leaves}(T)} R(t)
\end{equation}
$R(T)$ is the training error, $\text{Leaves}(T)$ gives the leaves of tree $T$,
$r(t) = 1-\max_k p(C_k)$ is the misclassification rate and $p(t)=n_t/N$ is the 
number of samples in node $n_t$ to total training samples N. Now the variation in cost complexity
is given by $R_\alpha(T-T_t) - R_\alpha(T)$, where $T$ is the complete tree, $T_t$ is the
subtree with root at node $t$, and a tree pruned at node $t$ would be $T-T_t$.
An ordering on the internal nodes for pruning is calculated by equating
the cost-complexity function $R_\alpha$ of pruned subtree $T-T_t$ to that of the branch at node $t$:

\begin{equation}
	g(t) = \frac{R(t)-R(T_t)}{|\text{Leaves}(T_t)|-1}
	\label{eqn:cost-complexity-param}
\end{equation}

\begin{figure}
	\centering
	\includegraphics[width=0.45\textwidth]{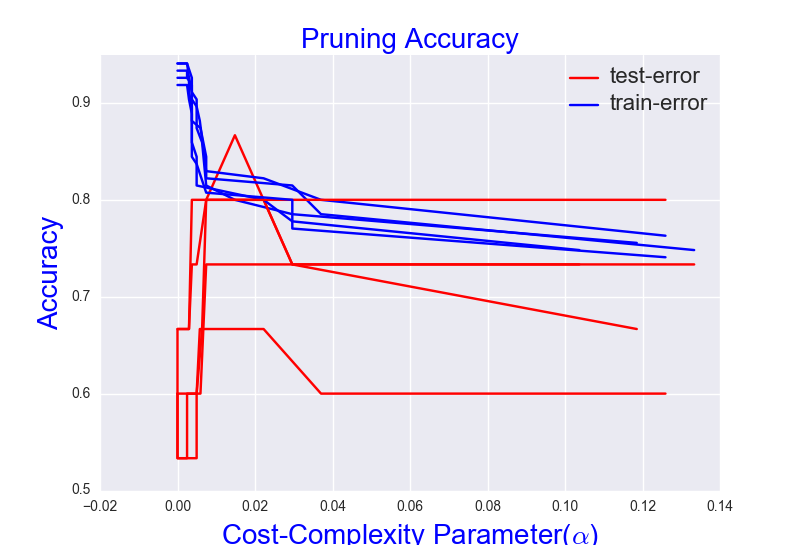}
	\includegraphics[width=0.45\textwidth]{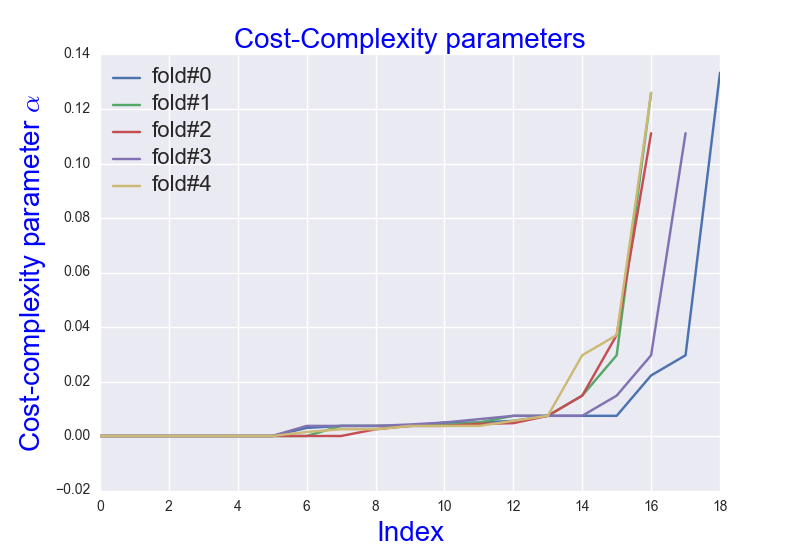}	
	\caption{Training error and averaged cross-validation error on 5 folds as cost-complexity parameter varies
	with its index. The index here refers to the number of subtrees.} 
		\label{fig:pruning_cross_val_error}
\end{figure}

The final step is to choose the weakest link to prune by calculating $\argmin g(t)$.
This calculation of $g(t)$ in equation (\ref{eqn:cost-complexity-param}) and then pruning
the weakest link is repeated until we are left with the root node. This provides 
a sequence of nested trees $\mathcal{T}$ and associated cost-complexity parameters $\mathcal{A}$.

In figure \ref{fig:pruning_cross_val_error} we plot the training 
error and test(cross-validation) error on 5 folds (usually 20 folds are used, 
this is only for visualization). We observe a deterioration in performance 
of both training and test errors. The small tree with 1 SE(standard error) 
of the cross-validation error is chosen as the optimal subtree. In our studies 
we use the simpler option which simply chooses the smallest tree with the smallest 
cross validation (CV) error.

\section{Out-of-Bag(OOB) cost complexity Pruning}

In Random forests, for each tree grown, $\frac{1}{e} N$ samples are not selected in
bootstrap, and are called out of bag (OOB) samples. The value $\frac{1}{e}$ refers to the 
probability of choosing an out-of-bag sample when $N\to \infty$. The OOB samples 
are used to provide an improved estimate of node probabilities and node
error rate in decision trees. They are also a good proxy for generalization
error in bagging predictors \cite{oobestimation_breiman_1996}.
OOB data is usually used to get an unbiased estimate of the classification
error as trees are added to the forest. 

The out-of-bag (OOB) error is the average error on the training set $Z$
predicted such that, samples from the OOB-set $Z \setminus Z_j$  
that do not belong to the set of trees $\{T_j\}$ are predicted with 
as an ensemble, using majority voting (using the sum of their class probabilities).

In our study (see figure \ref{fig:block_diag}) we use the OOB samples
corresponding to a given tree $T_j$ in the random forest ensemble, 
to calculate the optimal subtree $T^{\ast}_j$ by cross-validation. There are
two ways we propose to evaluate the optimal cost-complexity parameter, 
and thus the optimal subtree :
\begin{figure}
	\centering
	\includegraphics[width=\textwidth]{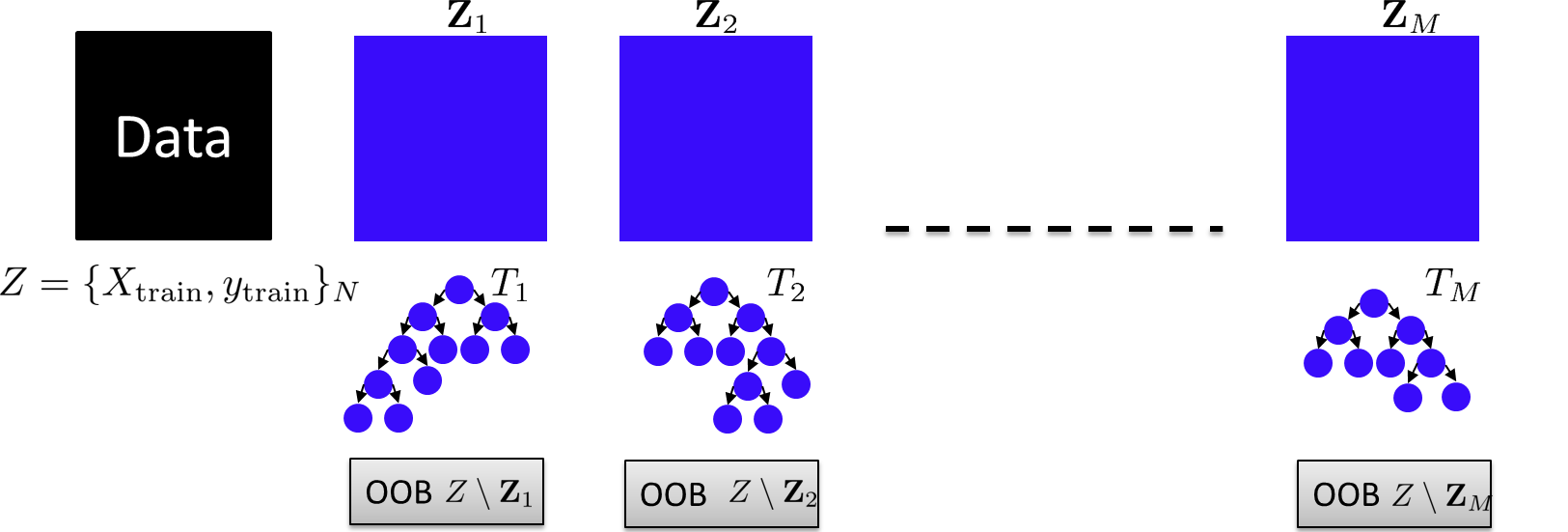}
	\caption{Figure shows the bagging procedure, OOB samples and its use for cost-complexity pruning on 
	each decision tree in the ensemble. There are two ways to choose the optimal subtree : 
	one uses the OOB samples as a cross-validation(CV) set to evaluate prediction error.
	The optimal subtree is the smallest subtree that minimizes the prediction error on the CV-set. }
	\label{fig:block_diag}
\end{figure}

\begin{itemize}
	\item Independent tree pruning : calculate the optimal subtree by evaluating
	\begin{equation}
		\mathcal{T}^\ast_j = \argmin_{\alpha \in \mathcal{A}_j} \mathbb{E} \bigg[\| Y_\text{OOB} - \mathcal{T}_j^{(\alpha)}(X^j_\text{OOB}) \|^2\bigg]
		\label{eqn:ind_tree_pruning}
	\end{equation}
	where $X^j_\text{OOB} = X_\text{train} \setminus X_j$, and $X_j$ being the samples used in the creation 
	of tree $j$.
	\item Global threshold pruning : calculate the optimal subtree by evaluating 
	\begin{equation}
		\{\mathcal{T}^\ast_j\}_{j=1}^M = \argmin_{\alpha \in \cup_j \mathcal{A}_j} \mathbb{E} \bigg[\| Y_\text{train} - \frac{1}{M} \sum_{j=1}^M 
		\mathcal{T}_j^{(\alpha)}(X^j_\text{OOB})\|^2 \bigg]
		\label{eqn:glob_tree_pruning}
	\end{equation}
	where the cross-validation uses the out-of-bag prediction error as to evaluate the optimal $\{\alpha_j\}$
	values. This basically considers a single threshold of cost-complexity parameters, which chooses
	a forest of subtrees for each threshold. The optimal threshold is calculated by cross-validating 
	over the training set.
\end{itemize}

The independent tree pruning and global threshold pruning are demonstrated in 
algorithmic form in figure \ref{alg:cc_prune_algos} as functions, BestTree\_byCrossValidation\_Tree
and BestTree\_byCrossValidation\_Forest. The main difference between them lies in the 
cross-validation samples and predictor (tree vs forest) used. 

\begin{algorithm}
  \caption{Creating random forest
    \label{alg:random_forest}}
  \begin{algorithmic}[1]
    \Require{$X_\text{train} \in \mathbb{R}^{N \times d}, Y_\text{train} \in \{C_k\}_1^K$, M-trees}
    \Statex
    \Function{CreateForest}{$\{T_i\}_{i=1}^M, X_\text{train}, y_\text{train}$}
      \For{$j \in$ [1, M]} 
	      \Let{$\mathbf{Z}_j$}{BootStrap($X_\text{train}, y_\text{train}$, N)}
	      \Let{$T_j$}{GrowDecisionTree($\mathbf{Z}_j$)}
	      \Comment{Recursively repeat : 1. select $m = \sqrt(d)$ variables 2. 
	      pick best split point among $m$ 3. Split node into daughter nodes. }
	      \Let{$Y_\text{pred}$}{$\argmax_{C_k} \frac{1}{M} \sum_{j=1}^M T_j(X_\text{test})$}
      \EndFor
      \Let{Error}{$\|Y_\text{pred}-Y_\text{test}\|^2$}
      \State \Return{$\{T_j\}_{j=1}^M$}
    \EndFunction
  \end{algorithmic}
\end{algorithm}

\begin{figure*}[htbp]
	\centering
	\includegraphics[width=\linewidth]{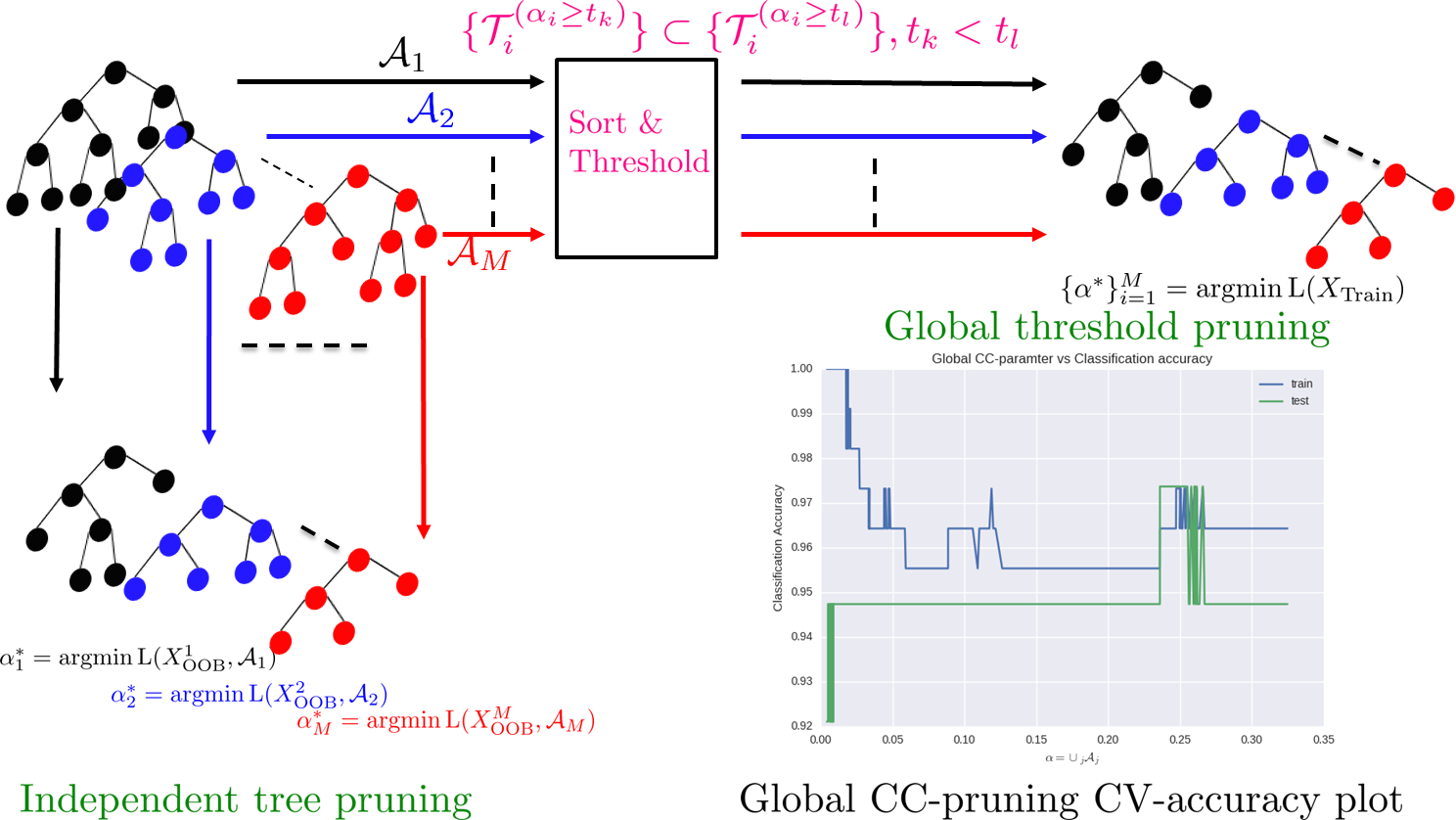}
	\caption{The two pruning methods from equations \ref{eqn:ind_tree_pruning} and \ref{eqn:glob_tree_pruning}
	are visually demonstrated. For the global method one can plot a training-vs-validation error plot 
	can be generated. This is very similar to figure \ref{fig:pruning_cross_val_error} for a single 
	decision tree.}
	\label{fig:global_overview}
\end{figure*}

\begin{algorithm}
  \caption{Cost complexity pruning on DTs
    \label{alg:pruning_trees_cost_complexity}}
  \begin{algorithmic}[1]
    \Require{$T_j$ a DT, $r$, re-substitution error, $p$ node occupancy proportion}
    \Statex
    \Function{CostComplexityPrune\_Tree}{$T_j, p, r$}
    \Let{$\mathcal{T}_j$}{$\emptyset$} \Let{$\mathcal{A}_j$}{$\emptyset$}
    \Comment{$\mathcal{T}_j$ for set of pruned trees \& cost-complexity parameter set $\mathcal{A}_j$.}
    \While{$T_j \neq \{0\}$ }
    \Comment{While tree is not pruned to root node.}
		\For{$ t \in T_i \setminus  \text{Leaves}(T_i) \setminus \{0\}$} 
		\Comment{For each internal node i.e. not a leaf/root}
			\Let{$R(t)$}{$r(t)*p(t)$}
			\Let{$R(T_t)$}{$\sum_{t \in  \text{Leaves}(n)} r(t)*p(t)$}
			\Let{$g(t)$}{$\frac{R(t)-R(T_t)}{|\text{Leaves}(n)|-1}$}
			\Let{$t_\alpha, \alpha$}{$\argmin g(t)$}
			\Comment{$\alpha^\ast$ is decided by cross-validation}
		\EndFor
		\Let{$\mathcal{T}_j$}{$\mathcal{T}_j \cup$ Prune($T_j, t_\alpha$)}
		\Let{$\mathcal{A}_j$}{$\mathcal{A}_j \cup \alpha$}
	\EndWhile
	\State \Return{$\mathcal{T}_j$, $\mathcal{A}_j$}
    \EndFunction
  \end{algorithmic}
\end{algorithm}

\begin{algorithm}
  \caption{Pruning DT ensembles
    \label{alg:pruning_forest_cost_complexity}}
  \begin{algorithmic}[1]
    \Require{$\{T_j\}_{j=1}^M$ are $M$ DTs from \textbf{RF}, \textbf{BT} or \textbf{ET}}
    \Statex
    \Function{CostComplexityPrune\_Forest}{$\{T_j\}_{i=1}^M, X_\text{train}, y_\text{train}$}
      \For{$i \in$ [1, n\_iter]} 
	      \For{$t \in \{T_j\}_{j=1}^M$} 
	      	  \Let{$\mathcal{T}_j, \mathcal{A}_j$}{CostComplexityPrune\_Tree($T_j, X_\text{OOB}^j$)}
	      	  \Let{$T^{\ast}_j$}{SmallestTree\_byCrossValidation($\mathcal{T}_j, \mathcal{A}_j, Z \setminus Z_j$)}
	      	  \Comment{$Z \setminus Z_j$ is replaced by $Z$ (the complete training set) to have a larger CV set(2nd algorithm).}
	      \EndFor
	      \Let{$T^{\ast\ast}_j$}{SmallestTree\_byCrossValidation\_Forest($\mathcal{T}, \mathcal{A}, \{Z \setminus Z_j\}$)}
	      \Let{$Y_\text{pred}$}{$\argmax_{C_k} \frac{1}{M} \sum_{j=1}^M T^{\ast}_j(X_\text{OOB}^j)$}
	      \Let{Error(i)}{$\|Y_\text{pred}-Y^\text{CV}_j\|^2$}
      \EndFor
      \State \Return{$T^{\ast}_j$}
    \EndFunction
  \end{algorithmic}
\end{algorithm}
The decision function of the decision tree (also denoted by the same symbol) $T_j$ 
would ideally map an input vector $\mathbf{x} \in \mathbb{R}^d$ to any of the $C_k$ classes
to be predicted. To perform the prediction for a given sample, we find nodes hit by
the sample until it reaches each leaf in the DT, and predict its majority vote. 	
The class-probability weighted majority vote across trees is frequently used since it 
provides a confidence score on the majority vote in each node across the different trees.

In algorithm (\ref{alg:pruning_forest_cost_complexity}) we evaluate the cost complexity pruning across the $M$
different trees $\{T_j\}$ in the ensemble, and obtain the optimal subtrees $\{T_J^\ast\}$ which minimize
the prediction error on the OOB sample set $Z \setminus Z_j$. 

One of the dangers of using the OOB set to evalute optimal subtrees individually, 
is that in small datasets the OOB samples might no more be representative of the 
original training samples distribution, and might produce large cross-validation errors. 
Though it remains to be studied whether using the OOB samples as a cross-validation set 
would effectively reduce the generalization error for the forest, even if we observe reasonable performance.

\begin{algorithm}
    \caption{Best subtree minimizing CV error on OOB-set  
    \label{alg:cc_prune_algos}}
    \begin{algorithmic}[1]
	    \Require{$\mathcal{T}_j$ set of nested subtrees of DT $T_j$, 
	    indexed by their cost-complexity parameter $\alpha \in \mathcal{A}_j$}
	    \Statex
	    \Function{{\bf \color{blue} BestTree\_byCrossValidation\_Tree}}{$\mathcal{T}_j, X_\text{OOB}^j$}
	    \For{$\alpha \in \mathcal{A}_j$} 
		    \Let{$Y_\text{pred}(\alpha)$}{$\argmax_{C_k} \mathcal{T}^{\alpha}_j(X_\text{OOB}^j)$}
		    \Let{CV-Error($\alpha$)}{$\|Y_\text{pred}(\alpha)-Y_\text{OOB}^j\|^2$}
		\EndFor
		\Let{$\alpha^{\ast}_j$}{$\argmin_{\alpha \in \mathcal{A}_j} \text{CV-Error}$}
	      \State \Return{$\mathcal{T}_j(\alpha^{\ast}_j)$}
	      \Comment{Returns the best subtree of $T_j$}
	    \EndFunction
  \end{algorithmic}

  \begin{algorithmic}[1]
	    \Require{$\{\mathcal{T}_j\}_{j=1}^M$ are $M$ DTs $T_j$ and their nested subtrees
	    indexed by their cost-complexity parameter $\alpha \in \mathcal{A}_j$ }
	    \Statex 
	    \Function{{\bf \color{blue} BestTree\_byCrossValidation\_Forest}}{$\{\mathcal{T}_j\}^M, \{\mathcal{A}_j\}^M, \{X_\text{OOB}^j\}^M$}
	    \Let{Unique\_alpha}{Unique($\bigcup_{j=1}^M \mathcal{A}_j$)}
	    \For{$a \in$ Unique\_alpha} 
	    	\Let{$\{\alpha_j\}$}{$\{\alpha \in \mathcal{A}_j | \alpha \leq a \}$}
		    \Let{$Y_\text{pred}(\{\alpha_j\})$}{$\argmax_{C_k} \frac{1}{M} \sum_{j=1}^M T^{\alpha}_j(X_\text{OOB}^j)$}
		    \Let{CV-Error($\{\alpha_j\}$)}{$\|Y_\text{pred}(\{\alpha_j\})-Y_\text{OOB}^j\|^2$}
		\EndFor
		\Let{$\{T^{\ast}_j\}$}{$\argmin_{\{\alpha_j\}} \text{CV-Error}(\{\alpha_j\})$}
	      \State \Return{$\{T^{\ast}_j\}$}
	      \Comment{Returns the $M$-best subtrees of $\{T_j\}_1^M$}
	    \EndFunction
  \end{algorithmic}

\end{algorithm}

\section{Experiments and evaluation}

Here we evaluate the Random Forest(RF), Extremely randomized tree(ExtraTrees, ET) 
and Bagged Trees (Bagger, BTs) models from scikit-learn on datasets from the 
UCI machine learning repository \cite{UCI_ML_repo_Lichman2013}. The data sets 
of different sizes are chosen. Datasets chosen were : Fisher's Iris 
(150 samples, 4 features, 3 classes), red wine (1599 samples, 11 features, 6 classes), 
white wine (4898 samples, 11 features, 6 classes), digits dataset (1797 samples, 64 features, 10 classes).
Code for the pruning experiments are available on github. 
\footnote{\href{https://github.com/beedotkiran/randomforestpruning-ismm-2017}{https://github.com/beedotkiran/randomforestpruning-ismm-2017}}
\begin{figure}[htbp]
	\centering
	\includegraphics[width=0.45\textwidth]{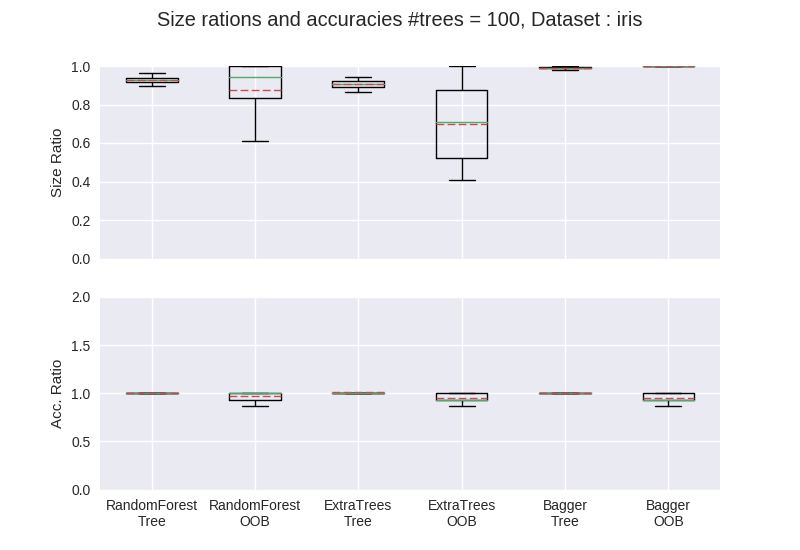}
	\includegraphics[width=0.45\textwidth]{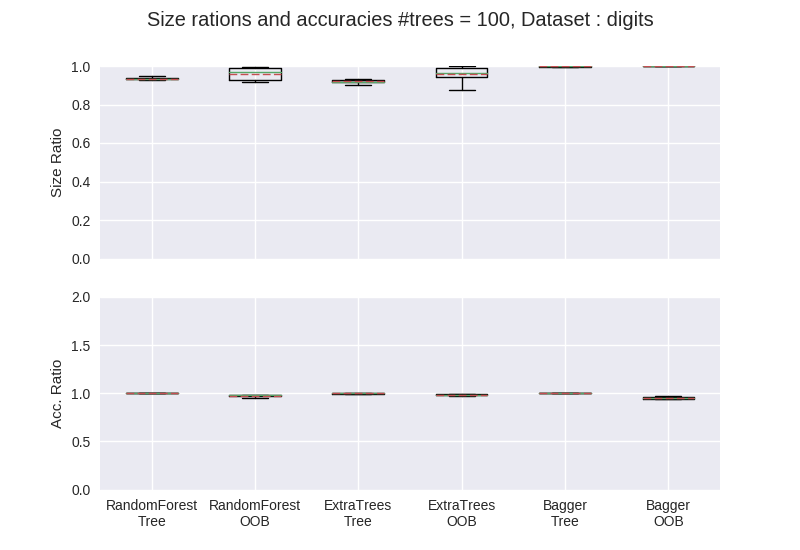}
	\includegraphics[width=0.45\textwidth]{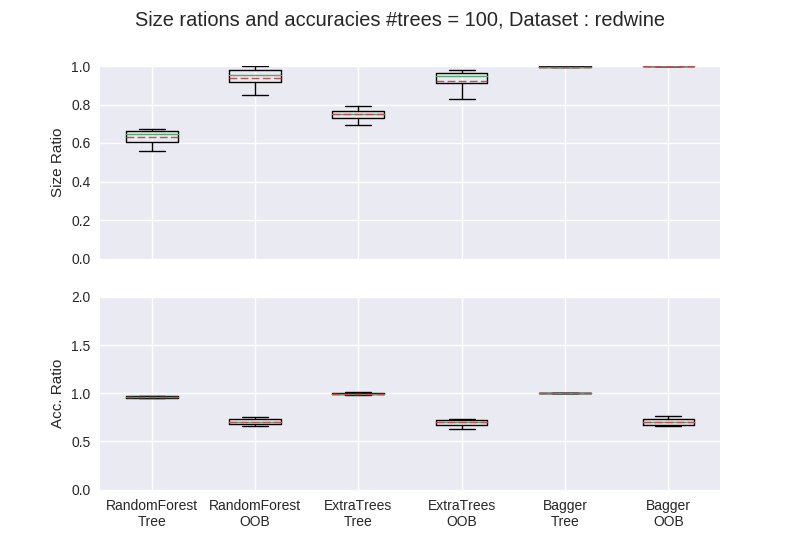}
	\includegraphics[width=0.45\textwidth]{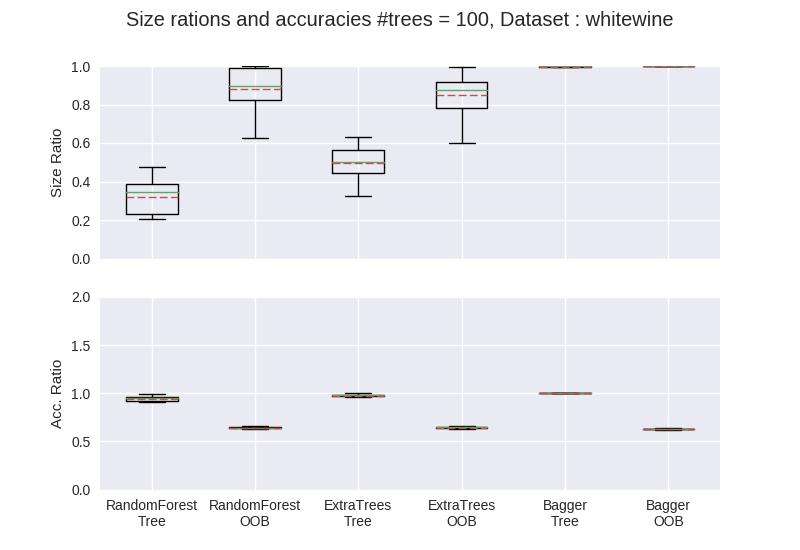}
	\caption{Performance of pruning on the IRIS dataset and digits dataset
	for the three models : RandomForest, Bagged trees and Extra Trees. For each
	model we show the performance measures for minimizing the tree prediction error
	and the forest OOB prediction error.}
	\label{fig:perf_plots}
\end{figure}

\begin{figure}[htbp]
	\centering
	\includegraphics[width=0.32\textwidth]{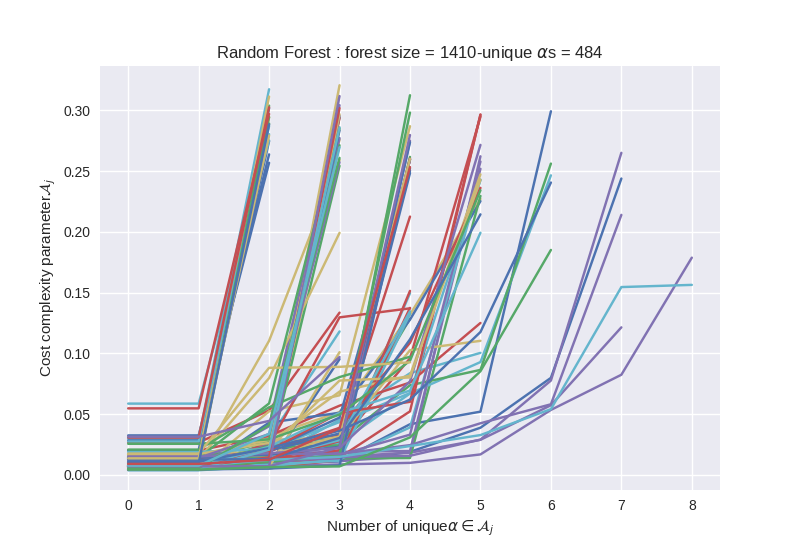}
	\includegraphics[width=0.32\textwidth]{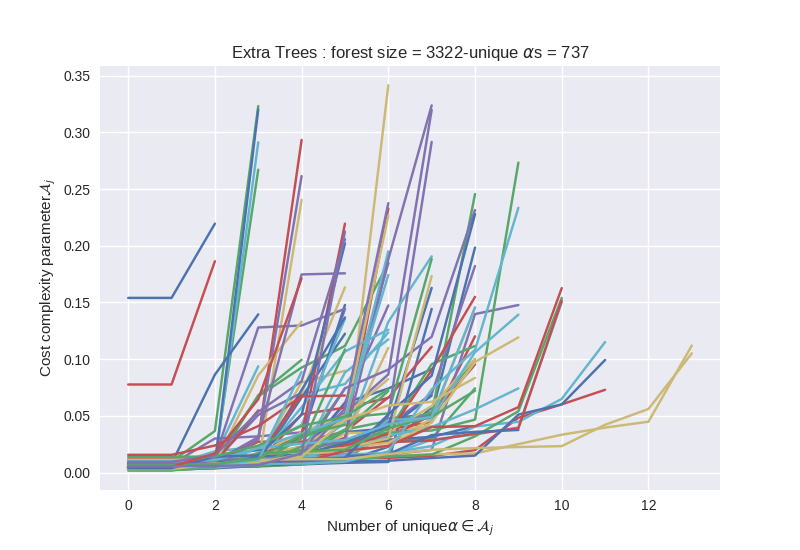}
	\includegraphics[width=0.32\textwidth]{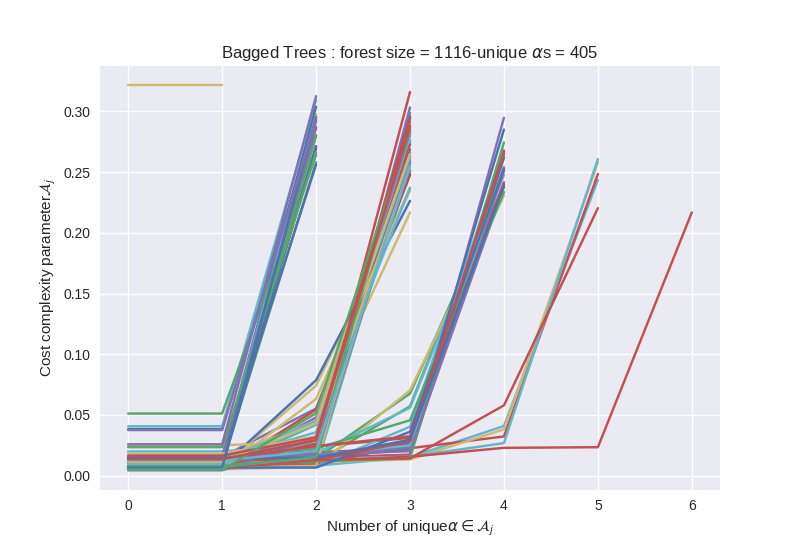}
	\caption{Plot of the cost-complexity parameters for 100-tree ensemble for RFs, ETs
	and BTs. The distribution of $\mathcal{A}_j$ across trees $j$ shows BTs constitute 
	of similar trees and thus contain similar cost complexity parameter values, while RFs
	and futhermore ETs have a larger range of parameter values, reflecting the known fact that these
	ensembles are further randomized. The consequence for pruning is that RFs and more so ETs
	on average produce subtrees of different depths, and achieve better prediction accuracy and size ratios 
	as compared to BTs.}
	\label{fig:OOB_vs_alpha}
\end{figure}

In figure \ref{fig:perf_plots} we demonstrate the effect of pruning RFs, 
BTs and ETs on the different datasets. We observe that random forests
an extra trees are often compressed by factors of 0.6 the original size,
while maintaining test accuracies, while this is not the case with BTs. 
To understand the effect of pruning we plot in figure \ref{fig:OOB_vs_alpha} 
the values $\mathcal{A}_j$ for the different trees in each of the 
ensembles. We observe that more randomization in RFs and ETs provide a larger
set of potential subtrees to cross-validate over.

Another important observation is seen in figure \ref{fig:global_overview}, 
as we prune the forest globally, the forest's accuracy on training set does 
not monotonically descend (as in the case of a decision tree). As we prune the 
forest, we could have a set of trees that improve their prediction while the others degrade. 

\section{Conclusions}

In this preliminary study of pruning of forests, we studied cost-complexity pruning of 
decision trees in bagged trees, random forest and extremely randomized trees. In our 
experiments we observe a reduction in the size of the forest which is dependent on the 
distribution of points in the dataset. ETs and RFs were shown to perform better than BTs, 
and were observed to provide a larger set of subtrees to cross-validate. This is the main
observation and contribution of the paper.

Our study shows that the out-of-bag samples can be a possible candidate to set
the cost-complexity parameter and thus an determine the best subtree for all DTs within ensemble. 
This combines the two ideas originally introduced by Breiman OOB estimates \cite{oobestimation_breiman_1996} and bagging 
predictors \cite{breiman2001randomForests}, while using the internal cross-validation OOB score of random forests
to set the optimal cost-complexity parameters for each tree.

The speed of calculation of the forest of subtrees is an issue. In the calculation of the 
forest of subtrees $\{\mathcal{T}\}_{j=1}^M$ we evaluate the predicitions at Unique($\cup_j \{\mathcal{A}_j\}$)
different values of the cost-complexity parameter, which represents the number of subtrees in the forest. 
In future work we propose to calculate the cost complexity parameter for the forest instead of individual trees. 

Though these performance results are marginal, the future scope and goal of this study is to identify the 
sources of over-fitting in random forests and reduce this by post-pruning. 
This idea might not be incompatible with smooth-spiked averaged decision 
function provided by random forests \cite{wyner2015explaining}.

\bibliographystyle{splncs03}
{\tiny
\bibliography{bibliography}
}
\end{document}